\documentclass[letterpaper]{article} 
\usepackage{aaai2026}  
\usepackage{times}  
\usepackage{helvet}  
\usepackage{courier}  
\usepackage[hyphens]{url}  
\usepackage{graphicx} 
\usepackage{natbib}  
\usepackage{caption} 
\usepackage{algorithm}
\usepackage{algorithmic}

\usepackage{amsmath}
\usepackage{amsthm}
\usepackage{amssymb}

\newtheorem{definition}{Definition}

\usepackage{booktabs}

\usepackage{newfloat}
\usepackage{listings}
\DeclareCaptionStyle{ruled}{labelfont=normalfont,labelsep=colon,strut=off} 
\floatstyle{ruled}
\newfloat{listing}{tb}{lst}{}
\floatname{listing}{Listing}
\title{DiffMM: Efficient Method for Accurate Noisy and Sparse Trajectory Map Matching via One Step Diffusion}
\author {
    Chenxu Han\textsuperscript{\rm 1},
    Sean Bin Yang\textsuperscript{\rm 2},
    Jilin Hu\textsuperscript{\rm 1,\rm 3}\thanks{Corresponding Author}
}
\affiliations {
    \textsuperscript{\rm 1}School of Data Science and Engineering, East China Normal University, Shanghai, China\\
    \textsuperscript{\rm 2}Department of Computer Science, Aalborg University, Aalborg, Denmark\\
    \textsuperscript{\rm 3}KLATASDS-MOE, East China Normal University, Shanghai, China\\
    cxhan@stu.ecnu.edu.cn, seany@cs.aau.dk, jlhu@dase.ecnu.edu.cn
}

\usepackage{bibentry}

\begin{document}

\maketitle

\begin{abstract}
  Map matching for sparse trajectories is a fundamental problem for many trajectory-based applications, e.g., traffic scheduling and traffic flow analysis. Existing methods for map matching are generally based on Hidden Markov Model (HMM) or encoder-decoder framework. However, these methods continue to face significant challenges when handling noisy or sparsely sampled GPS trajectories. To address these limitations, we propose \emph{DiffMM}, an encoder–diffusion-based map matching framework that produces effective yet efficient matching results through a one-step diffusion process. We first introduce a road segment-aware trajectory encoder that jointly embeds the input trajectory and its surrounding candidate road segments into a shared latent space through an attention mechanism. Next, we propose a one step diffusion method to realize map matching through a shortcut model by leveraging the joint embedding of the trajectory and candidate road segments as conditioning context. We conduct extensive experiments on large-scale trajectory datasets, demonstrating that our approach consistently outperforms state-of-the-art map matching methods in terms of both accuracy and efficiency, particularly for sparse trajectories and complex road network topologies. 
\end{abstract}

\begin{links}
    \link{Code}{https://github.com/decisionintelligence/DiffMM}
\end{links}

\section{Introduction}

Map matching is a fundamental component in map service that aligns the vehicle or human trajectory records with the underlying the road network. It is of vital importance in many trajectory-based applications, including vehicle navigation~\cite{navi}, traffic flow prediction~\cite{tfb, duet}, traffic scheduling~\cite{trafsch}, route optimization~\cite{routeopt}. For instance, in services like Google Maps, accurate map matching is crucial for delivering reliable navigation by precisely localizing users and estimating real-time traffic conditions on the road network.

\begin{figure}
    \centering
    \includegraphics[width=\linewidth]{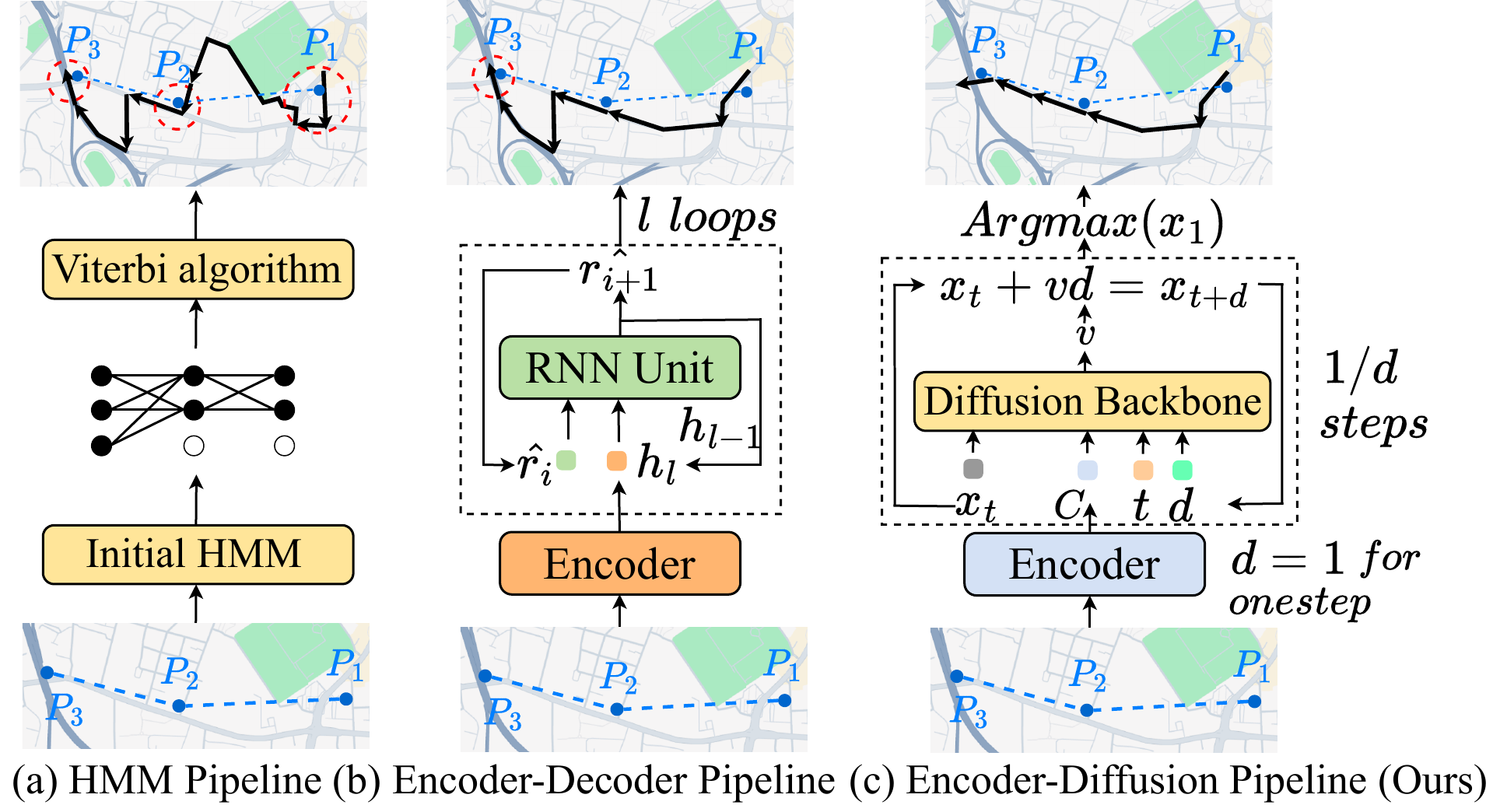}
        \caption{Comparison of Different Map Matching Pipelines. The black arrows represent the result matched road segments and the red dashed circles highlight the wrong result.}
    \label{fig:example}
\end{figure}

 Map matching has been extensively studied over the past years, resulting in a wide range of proposed methods. These approaches can generally be categorized into two main groups: non-learning-based methods~\cite{hmm, var1, var2, var3} and learning-based methods \cite{deepmm, gmm, rntraj}.
One of the most widely used non-learning-based approaches for map matching is the Hidden Markov Model (HMM)~\cite{hmm}, which models the sequence of observed GPS points as emissions from a hidden sequence of road segments, effectively capturing both spatial proximity and temporal continuity. 
In contrast, learning-based methods have gained increasing attention in recent years. Several end-to-end approaches~\cite{deepmm, gmm, rntraj} have been proposed, which consistently outperform traditional HMM-based methods. These approaches are primarily based on the sequence-to-sequence (Seq2Seq) paradigm and adopt an encoder-decoder framework to model the mapping from noisy GPS trajectories to the corresponding road segment sequences. 

Taking Figure \ref{fig:example} as as example, we observe the following: (1) As shown in Figure~\ref{fig:example}(a), traditional HMM-based methods are highly sensitive to noisy and sparsely sampled trajectory data, leading to degraded matching performance, particularly in the regions highlighted by the red dashed circle; (2) As shown in Figure~\ref{fig:example}(b), although learning-based methods demonstrate improved performance over non-learning-based approaches, their matching accuracy is still adversely affected by sparse trajectory data.
Moreover, the autoregressive nature of RNN-based decoders introduces inefficiencies and tends to accumulate errors during the decoding process, ultimately leading to suboptimal results.
To this end, despite substantial advancements, both non-learning-based and learning-based map matching methods continue to face significant challenges when handling noisy or sparsely sampled GPS trajectories.

Specifically, we highlight the following two major limitations of existing methods:

 \begin{itemize}
     \item \textbf{Sensitive to noise in trajectory data.} Due to environmental complexity and limitations of positioning devices, GPS data often exhibit significant drift, causing recorded points to deviate from their true locations. Such noise severely impacts the accuracy of map matching, particularly for distance-based methods such as HMM, which rely heavily on spatial proximity for candidate selection and state transitions.
     \item \textbf{Degraded performance on sparse trajectories.} To reduce storage costs, many map services adopt low-frequency sampling strategies, resulting in sparsely recorded GPS trajectories. While this approach is storage-efficient, it poses significant challenges for map matching. The limited number of observations hampers the ability to capture temporal and spatial continuity between consecutive points, making it difficult to infer accurate road segment sequences. Consequently, the matching accuracy is often significantly degraded when dealing with sparse trajectories.
 \end{itemize}

To address these limitations, we propose \emph{DiffMM}, an encoder–diffusion-based map matching framework that produces effective yet efficient matching results through a one-step diffusion process. 
Specifically, we first introduce a road segment-aware trajectory encoder that jointly embeds the input trajectory and its surrounding candidate road segments into a shared latent space. This is achieved through an attention mechanism that captures the interactions between trajectory points and road segments, allowing the model to focus on spatially and contextually relevant features. To construct the candidate set, we adopt a $\delta$-meter radius strategy, which selects all road segments within a predefined distance $\delta$ from each GPS point, ensuring that the candidate segments are both spatially meaningful and computationally tractable.
Next, we propose a distribution-based map matching model that leverages the joint embedding of the trajectory and candidate road segments as conditioning context. Based on this representation, we formulate the map matching task as a one-step diffusion process, where the target road segment distribution is recovered via a denoising mechanism applied to Gaussian noise. Specifically, we adopt a shortcut model to approximate the conditional distribution in a single denoising step, significantly enhancing the efficiency of both training and inference.
We conduct extensive experiments on large-scale trajectory datasets to evaluate the effectiveness of \emph{DiffMM}. The results demonstrate that our approach consistently outperforms state-of-the-art map matching methods in terms of both accuracy and efficiency, particularly under challenging conditions involving highly sparse trajectories and complex road network topologies.
We summarize our contributions as follows:

 \begin{itemize}
     \item We propose a novel one-step diffusion-based map matching framework, \emph{DiffMM}. To the best of our knowledge, we are the first to model Map Matching through the conditional distribution, which is within diffusion paradigm and allows us to leverage the information of trajectory and road network. 
     \item We propose a road segment-aware trajectory encoder that jointly embeds the input trajectory and its surrounding candidate road segments into a shared latent space through an attention mechanism. 
     \item We propose a one step diffusion method to realize map matching through a shortcut model by leveraging the joint embedding of the trajectory and candidate road segments as conditioning context.
     \item We conduct extensive experiments on large-scale trajectory datasets, demonstrating that our approach consistently outperforms state-of-the-art map matching methods in terms of both accuracy and efficiency, particularly for sparse trajectories and complex road network topologies. 
 \end{itemize}

\section{Related Work}

\subsection{Map Matching}

 There is a plethora of studies focus on map matching. The classic study~\cite{hmm} leverages Hidden Markov Model to find the most likely matched road segments, and it have many variants~\cite{var1, var2, var3, fmm}.~\cite{rinfer} present a history-based route infer system.~\cite{lowmm} leverage the road network and the temporal features of trajectories to construct a candidate graph. FMM~\cite{fmm} apply some acceleration mechanisms on HMM. Recent trend is to extract patterns from historical trajectories for map matching. DeepMM~\cite{deepmm} is an end-to-end deep learning method with data augmentation. GraphMM~\cite{gmm} is a graph-based method which incorporate graph neural networks to leverage both intra-trajectory and inter-trajectory correlation for map matching. RNTrajRec~\cite{rntraj} is an approach to solve both map matching and trajectory recovery by considering the road network structure. Different from the existing methods, we formulate map matching as a problem of learning a conditional distribution and solve it by DiffMM. 

\subsection{Diffusion Models}

 Diffusion models~\cite{ddpm, ddpm2, ddpm3, shortcut} are a class of generative models. There are rich studies on diffusion models in various domains, and the diffusion models have been widely used for many applications due to their powerful generative capabilities. These applications include image generation~\cite{img1, img2, img3, img4}, time series prediction and imputation~\cite{ts1, ts2}, text generation~\cite{text1, text2}, audio generation~\cite{audio1, audio2, audio3}, path planning~\cite{pp}, trajectory generation~\cite{trajgen} and spatio-temporal point processes~\cite{dstpp}. In this paper, we are the fist to introduce the diffusion model to map matching. 

\section{Preliminaries}

\subsection{Definitions}

\begin{definition}[Road Network]
 The road network is represented as a directed graph $G=(V, E)$, where $V$ is a set of nodes and $E$ is a set of directed edges. Each node $v_i \in V$ represents an intersection or a end point of a road. Each edge $e = (u, v) \in E$ is a road segment from the entrance node $u$ to the exit node $v$. $|V|$ denotes the number of intersections and $|E|$ denotes the number of road segments. 
\end{definition}

\begin{definition}[Trajectory]
 A trajectory $T$ is defined as a sequence of GPS points with timestamps, i.e., $T = (p_1, p_2, ..., p_l)$, where $l$ is the trajectory length. Each GPS point $p_i=(lat, lng, t) \in T$ consisting of latitude $lat$, longitude $lng$ and timestamp $t$. 
\end{definition}

\subsection{Diffusion and Shortcut}

 Diffusion~\cite{ddpm} and flow-matching~\cite{flow} models approach the generative modeling problem by learning an ordinary differential equation~(ODE) that transforms noise into data. We consider flow-matching as a special case of diffusion modeling~\cite{diffflow} and use the terms interchangeably. Recently, a new family of denoising generative models called shortcut models overcomes the large number of sampling steps required by diffusion and flow-matching models by introducing desired step size $d$ into the models. We define $x_t$ as a linear interpolation between a data point $x_1 \sim \mathcal{D}$ and a noise point $x_0 \sim \mathcal{N}(0, \mathbb{I})$ of the same dimensionality. The velocity $v_t$ is the direction from the noise to the data point:

\begin{equation}
    x_t=(1-t)x_0+tx_1 \quad \text{and} \quad v_t=x_1-x_0.
\end{equation}

 Flow-matching models learn a neural network to estimate the expected value $\overline{v}_t=\mathbb{E}[v_t | x_t]$ that averages over all possible velocities at $x_t$. Conditioning on step size $d$, shortcut models refer to the normalized direction from $x_t$ towards the correct next point $x'_{t+d}$ as $s(x_t, t, d)$:

\begin{equation}
    x'_{t+d} = x_t + s(x_t, t, d)d.
\end{equation}

 At $d \xrightarrow{} 0$, shortcut is equivalent to the flow. Shortcut models have an inherent self-consistency property, namely that one step equals two consecutive steps of half the size:

\begin{equation}
    s(x_t, t, 2d) = s(x_t, t, d)/2 + s(x'_{t+d}, t+d, d)/2.
\end{equation}

 This allows shortcut models to be trained using self-consistency targets for $d>0$ and using the flow-matching loss as a base case for $d=0$. Therefore the shortcut models can be optimized by the combined shortcut model target:

\begin{equation}
\begin{aligned}
 & \mathbb{E}_{x_0, x_1}[\| s_\theta(x_t, t, 0) - (x_1 - x_ 0) \|^2 + \| s_\theta(x_t, t, 2d) - s_{t} \|^2],
\end{aligned}
\end{equation}
\noindent
where $s_{t}=s_\theta(x_t, t, d)/2 + s_\theta(x'_{t+d}, t+d, d)/2$ and $  x'_{t+d}=x_t + s_\theta(x_t, t, d)d$.

\subsection{Problem Statement} 

Given a road network $G=(V,E)$ and a trajectory $T=(p_1,...,p_l)$, map matching is to map $(p_1,...,p_l)$ to a sequence of edges in $G$, denoted as $R=(e_1,...,e_l)$, where $e_i \in E$. We refer to $R$ as the route, which is essentially a sequence of matched road segments.

\section{Methodology}

\begin{figure*}[t]
\centering
\hspace{-0.7cm}
\includegraphics[width=0.86\textwidth]{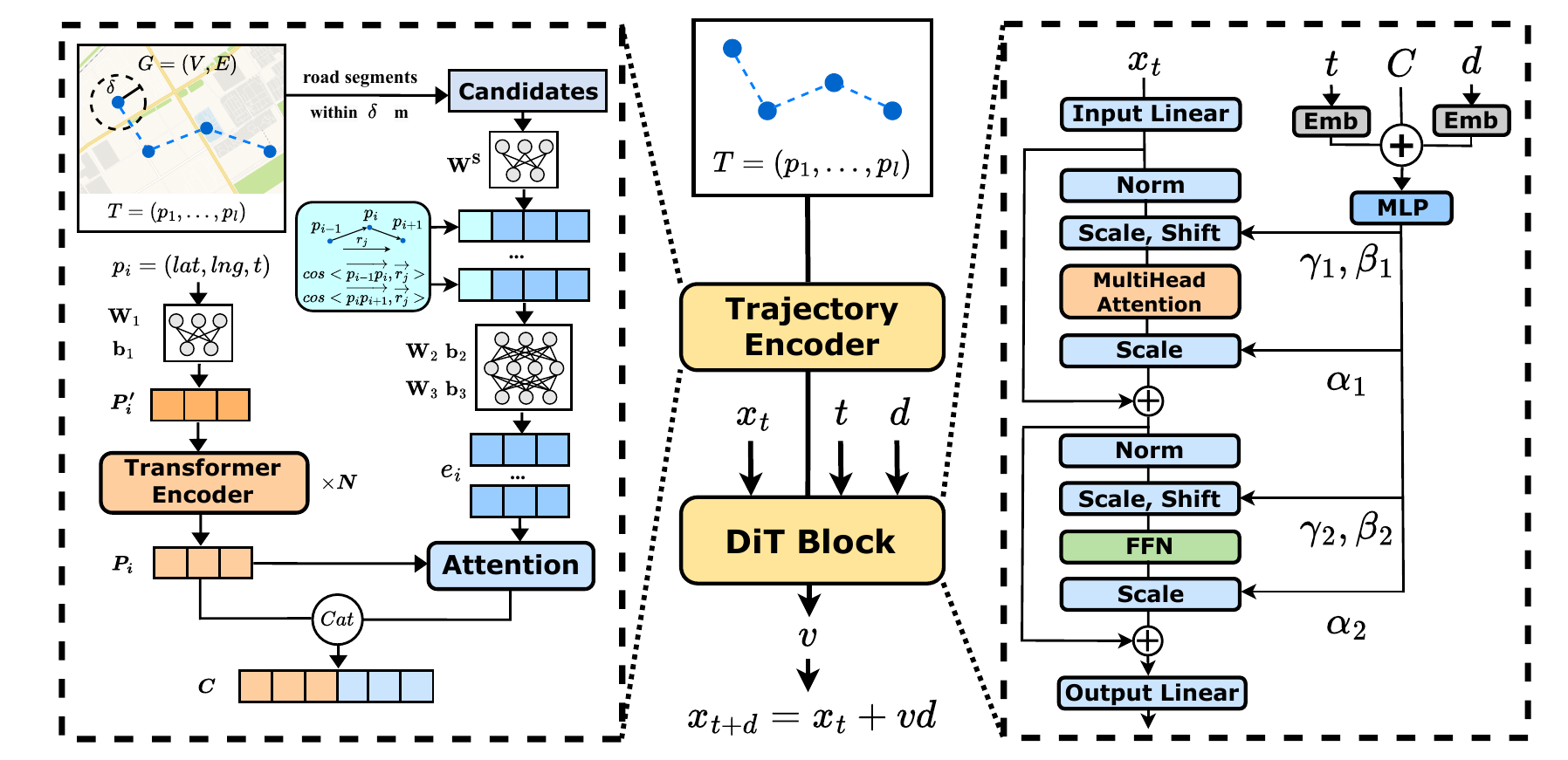}
\caption{The overview of DiffMM.}
\label{diffmm}
\end{figure*}

 Figure \ref{diffmm} illustrates the overall framework of DiffMM that contains two key modules: the trajectory encoder and the backbone DiT block~\cite{dit}. The Trajectory Encoder learns an effective representation $\boldsymbol{\mathit{C}}$ of the input trajectory by point representation and segment representation. To address the limitation of sparse trajectory, point representation adopts Transformer encoder to obtain the sequential dependencies of trajectory. To reduce the harm of noise, segment representation considers the nearby segments as candidates and utilizes attention mechanism to fusion the candidate segment embeddings. Then $\boldsymbol{\mathit{C}}$ serves as one of the conditions for the shortcut model in the denoising process. We choose DiT Block as the backbone of shortcut model to further utilize the sequential dependencies of trajectory, which is helpful for alleviating the problem of sparse trajectory. We first introduce the details of the trajectory encoder. Then we formulate the shortcut for DiffMM and present the architecture of the DiT Block. Finally, we describe how DiffMM is trained and the inference process of it.

\subsection{Trajectory Encoder}

To address the limitations mentioned above, we design an attention-based Trajectory Encoder which encodes raw GPS data, sequential information, directional information, and road network information together to obtain an effective trajectory representation for shortcut model. The input of the encoder is the trajectory $T=(p_1,...,p_l)$ and the road network $G=(V,E)$ and the output is $\boldsymbol{\mathit{C}}\in \mathbb{R}^{l\times d_{cond}}$, where $l$ is the length of the trajectory and $d_{cond}$ is the dimension of the condition to be used in DiT block.

  \subsubsection{Point Representation.}
  To address the limitation of low sampling rate, Trajectory encoder obtain the representation of the raw GPS sequence~(trajectory points) utilizing the raw GPS data and sequential information. Initially, each point $p_i$ can be represented as a three-element-vector $\boldsymbol{p}_i^{(0)}$ containing its min-max normalized latitude, longitude and timestamp. Then $\boldsymbol{p}_i^{(0)}$ is fed into a fully connected network to get $\boldsymbol{p}_i^{(1)}$ by $\boldsymbol{p}_i^{(1)}= \boldsymbol{p}_i^{(0)}\mathbf{W}_1 + \mathbf{b}_1$, where $\mathbf{W}_1 \in \mathbb{R}^{3\times d_{emb}}$ and $\mathbf{b}_1 \in \mathbb{R}^{d_{emb}}$ are the learnable parameters. 

  After obtaining the embeddings of all GPS points $\boldsymbol{P'}=[\boldsymbol{p}_1^{(1)},...,\boldsymbol{p}_l^{(1)}]$ in the trajectory, we adopt Transformer encoder~\cite{transformer} to capture the sequential dependencies in the trajectory:

  \begin{equation} \label{pointtrans}
  \boldsymbol{P}=TransEncoder(\boldsymbol{P'}).
  \end{equation}

  A Transformer encoder layer consists of two sub-layers: a multi-head self-attention and a feed-forward network~(FFN). Multi-head attention is given by:
  \begin{equation}
  \begin{aligned}
  & MultiHeadAttn(\mathbf{Q},\mathbf{K},\mathbf{V}) = concat(head_1,...head_h) \mathbf{W}^O,\\
  & head_i= Attn(\mathbf{Q}\mathbf{W}_i^Q, \mathbf{K}\mathbf{W}_i^K, \mathbf{V}\mathbf{W}_i^V),\\
  & Attn(\mathbf{Q},\mathbf{K},\mathbf{V}) = softmax(\frac{\mathbf{Q} \mathbf{K}^T}{\sqrt{d}})\mathbf{V},
  \end{aligned}
  \end{equation}
  where $\mathbf{Q}$, $\mathbf{K}$ and $\mathbf{V}$ are the query, key and value matrix, $\mathbf{W}_i^Q,\mathbf{W}_i^K,\mathbf{W}_i^V$ are parameters for the i-th head, $h$ is the number of attention heads, $\mathbf{W}^O$ is parameter for output. 

  An FFN is two-layer MLP with ReLU activation:

  \begin{equation}
  FFN(\mathbf{X}) = ReLU(\mathbf{X}\mathbf{W} + \mathbf{b}) \mathbf{W'} + \mathbf{b'},
  \end{equation}
  where $\mathbf{W},\mathbf{b},\mathbf{W'}$ and $\mathbf{b'}$ are learnable parameters.

  Let $\mathbf{X}$ be the input, a Transformer layer applies $MultiHeadAttn$ to $\mathbf{X}$ itself followed by an $FFN$, both with residual connection and layer normalization:

  \begin{equation}
  \begin{aligned}
  & TransEncoder(\mathbf{X}) = LayerNorm(\mathbf{X'} + FFN(\mathbf{X}')),\\
  & \mathbf{X'} = LayerNorm(\mathbf{X} + MultiHeadAttn(\mathbf{X},\mathbf{X},\mathbf{X})).
  \end{aligned}
  \end{equation}

  \subsubsection{Segment Representation.} 
  To reduce the effects of noisy records, we locate the road segments within $\delta$ meters from $p_i$ via R-tree~\cite{rtree} data structure as candidate segments for each GPS point $p_i$ in the trajectory. We then incorporate the candidate segments as road network context to build more informative representation of the trajectory. 
  
  For each segment $r_{ij}$ of the candidate segments of $p_i$, we first apply a fully-connected layer to embed the segment $r_{ij}$ to a high dimension space, i.e., $\mathbb{R}^{d_{emb}}$.

  \begin{equation}
  \mathbf{e}^{(0)}_{r_{ij}} = \mathbf{1}_{r_{ij}} \mathbf{W^{S}},
  \end{equation}
  where $\mathbf{W^{S}} \in \mathbb{R}^{|E|\times d_{emb}}$ is learnable parameter, $|E|$ represents the number of road segments, $\mathbf{1}_{r_{ij}} \in \{0, 1\}^{|E|}$ is the one-hot vector representing $r_{ij}$ with all elements are 0 except 1 at the position corresponding to the segment id of $r_{ij}$.

  To integrate the direction information, we view each segment $r_{ij}$ as a vector and calculate two cosine similarity: 1) its similarity with the vector from $p_{i-1}$ to $p_i$, 2) its similarity with the vector from $p_{i}$ to $p_{i+1}$. Also, we calculate the distance between $p_i$ and its projection on the road segment $r_{ij}$, and concatenate these values with $\mathbf{e}^{(0)}_{r_{ij}}$ to get $\mathbf{e}^{(1)}_{r_{ij}} \in \mathbb{R}^{d_{emb}+3}$. We produce the final segment embedding $\mathbf{e}_{r_{ij}}$ for $r_{ij}$ through a Multi-layer Perceptron~(MLP):

  \begin{equation}
  \mathbf{e}_{r_{ij}} = ReLU(\mathbf{e}^{(1)}_{r_{ij}} \mathbf{W}_2 + \mathbf{b}_2) \mathbf{W}_3 + \mathbf{b}_3,
  \end{equation}
  where $\mathbf{W}_2 \in \mathbb{R}^{(d_{emb}+4)\times d_{emb}}$, $\mathbf{b}_2 \in \mathbb{R}^{d_{emb}}$, $\mathbf{W}_3 \in \mathbb{R}^{d_{emb}\times d_{emb}}$ and $\mathbf{b}_3 \in \mathbb{R}^{d_{emb}}$ are learnable parameters.

  We have obtained the candidate segment embeddings $[\mathit{R}_1, ..., \mathit{R_l}]$ for the trajectory, where $\mathit{R}_i \in \mathbb{R}^{n\times d_{emb}}$, $n$ is the number of candidate segments of $p_i$. However, the number of road segments within $\delta$ meters from each $p_i$ could be very different. To address this, we design an attention mechanism to fuse the candidate segments embeddings reasonably to obtain the segment representation. Intuitively, the attention weight for the candidate segment $r_{ij}$ are relative to the GPS point $p_i$ and itself. Therefore, the fusion of the candidate segments embeddings of $p_i$ are as follows:

  \begin{equation}
  \begin{aligned}
  & \mu_{j, i} = ReLU(concat(\boldsymbol{P}[i],e_{j})\mathbf{W}_4 + \mathbf{b}_4)\mathbf{W}_5+\mathbf{b}_5,\\
  & w_{j, i} = \frac{exp(\mu_{j, i})}{\sum_{\forall s\in \mathit{C}_i}exp(\mu_{s, i})},\\
  & f_i = \sum_{\forall j\in \mathit{C}_i} w_{j, i}\cdot e_j,
  \end{aligned}
  \end{equation}
  where $\mathbf{W}_4 \in \mathbb{R}^{2d_{emb}\times d_a}$, $\mathbf{b}_4\in \mathbb{R}^{d_a}$, $\mathbf{W}_5 \in \mathbb{R}^{d_a\times 1}$ and $\mathbf{b}_5 \in \mathbb{R}^{1}$ are learnable parameters, $\boldsymbol{P}[i]$ represents the point representation of $p_i$, $\mathit{C}_i$ is the candidate segments of $p_i$ and $e_j$ represents the embedding of road segment $j$. 

  The final trajectory representation $\boldsymbol{\mathit{C}}$ is the concatenation of point representation and segment representation for each point in the trajectory, i.e., $\boldsymbol{\mathit{C}}=[\mathit{c}_1,...,\mathit{c}_l]$, $\mathit{c}_i=Concat(\boldsymbol{P}[i], f_i)$, $\boldsymbol{\mathit{C}}\in \mathbb{R}^{l\times d_{cond}}$, $d_{cond}=2d_{emb}$, and $l$ is the length of the trajectory.

\subsection{DiffMM Shortcut Model}

 To further address the problem of low sampling ratio, we use a shortcut model~\cite{shortcut} to learn the distribution conditioned on the trajectory representation $\boldsymbol{\mathit{C}}$ generated by trajectory encoder. Let $x_1$ represents the flow target given the trajectory $\mathit{T}$, $x_1 \in \mathbb{R}^{l\times |E|}$ where $|E|$ is the number of road segments and $l$ is the length of the trajectory. $x_1[i,j]$ is set to 1 if $p_i$'s ground truth matched segment is $j$ otherwise 0. Let $x_0 \in \mathbb{R}^{l\times |E|} \sim \mathcal{N}(0,\mathbb{I})$ represents the start Gaussian noise, value of $x_0[i][j]$ represents the probability that \textit{i-th} point match on segment \textit{j}. We define $x_t$ as a linear interpolation between $x_1$ and $x_0$, the shortcut with condition embedding $\boldsymbol{\mathit{C}}$ generated from the original trajectory as follows: 

\begin{equation}
    x_{t+d} = x_t + s(x_t, t, d, C)d.
\end{equation}

 It also has the self-consistency property:
 
\begin{equation}
    s(x_t, t, 2d, C) = s(x_t, t, d, C)/2 + s(x_{t+d}, t+d, d, C)/2.
\end{equation}

 In this way, we incorporate $C$ into the shortcut model to obtain the conditional distribution. Also, DiffMM can be trained in the shortcut's manner. We choose DiT Block~\cite{dit} as the backbone of the shortcut, it first maps input $x_t$ to the model dimension $d_{model}$ through a linear and embeds the time $t$ and the desired step size $d$ by sinusoidal embedding as follows:

\begin{equation}
SinEmb(t)= \left\{
\begin{aligned}
    & cos(t/10000^{\frac{j-1}{d}}) & if \  j \  is \ odd \\
    & sin(t/10000^{\frac{j-1}{d}}) & if \  j \  is \ even ,
\end{aligned}
\right.
\end{equation}
where $d$ denotes the embedding dimension.

 The condition for DiT is $cond=C+SinEmb(t)+SinEmb(d)$, and the multi-head attention part is calculated by the following steps:

\begin{equation}
\begin{aligned}
    & \alpha_1,\beta_1, \gamma_1=MLP(cond),\\
    & x_t' = \gamma_1 Norm(x_t)+\beta_1,\\
    & x = x_t + \alpha_1 MultiHeadAttn(x_t',x_t',x_t').
\end{aligned}
\end{equation}

 The FFN part is similar as the multi-head attention part, then the final output is mapped to the original dimension by the output linear for the shortcut process. 

\subsection{Training and Inference}

\begin{algorithm}[tb]
\caption{DiffMM Training}
\label{train}
\textbf{Input}: Trajectories with matched-segments $D$, road network $G=(V, E)$
\begin{algorithmic}[1] 
\WHILE{not converged}
\STATE $T, x_1 \sim D,x_0\sim \mathcal{N}(0, \mathbb{I}), (d,t)\sim p(d, t)$
\STATE $x_t \xleftarrow{} (1-t)x_0 + tx_1$
\STATE $C=TrajEncoder_\phi(T,G)$
\FOR{first k batches}
\STATE $s_{target} \xleftarrow{} x_1-x_0$
\STATE $d \xleftarrow{} 0$
\ENDFOR

\FOR{other batches}
\STATE $s_t \xleftarrow{} s_\theta(x_t, t, d, C)$
\STATE $x_{t+d} \xleftarrow{} x_t+s_td$
\STATE $s_{t+d} \xleftarrow{} s_\theta(x_{t+d},t+d,d,C)$
\STATE $s_{target} \xleftarrow{} (s_t+s_{t+d})/2$
\ENDFOR
\STATE Take gradient descent on $\phi$ and $\theta$ by Equation (\ref{lossall})
\ENDWHILE
\end{algorithmic}
\end{algorithm}

 \subsubsection{Training.} 
 DiffMM is trained by two loss functions: shortcut loss and cross entropy loss. The shortcut loss is defined as follows:

 \begin{equation}
     \mathcal{L}_{st}=\| s_\theta(x_t, t, 2d, C) - s_{target} \|^2.
 \end{equation}
 It is flow-matching target when $d=0$, otherwise it is the self-consistency target. 

 We also add an auxiliary cross entropy loss for training:

 \begin{equation}
     \mathcal{L}_{ce}=CrossEntropy(x_1, x_t+s_t),
 \end{equation}
 where $x_1$ represents the target segments.

 The total loss of DiffMM is:
 \begin{equation}
 \label{lossall}
 \mathcal{L}=\mathcal{L}_{st}+\mathcal{L}_{ce}.
 \end{equation}
 We train the overall framework~(trajectory encoder and DiT Block) in an end-to-end manner, the pseudo code of the training is shown in Algorithm~\ref{train}.

\begin{algorithm}[tb]
\caption{DiffMM Inference}
\label{infer}
\textbf{Input}: Trajectory $T$, road network $G$, sample step $M$ \\
\textbf{Output}: Matched segments $R$
\begin{algorithmic}[1]
\STATE $x \sim \mathcal{N}(0,\mathbb{I})$, $d \xleftarrow{} 1/M$, $t \xleftarrow{} 0$
\STATE $C=TrajEncoder_\phi(T, G)$
\FOR{$i \in [0,...,M-1]$}
\STATE $x \xleftarrow{} x+s_\theta(x,t,d,C)d$
\STATE $t \xleftarrow{} t+d$
\ENDFOR
\STATE $R \xleftarrow{} []$
\FOR{$i \in [0,...,T.length]$}
\STATE $R.append(Argmax(x[i]))$
\ENDFOR
\STATE \textbf{return} $R$
\end{algorithmic}
\end{algorithm}

 \subsubsection{Inference.}
 During inference, DiffMM first generates condition embedding by Trajectory Encoder, then obtain the conditional distribution $x$ by shortcut model. We treat the most possible matched segment of each point $i$ as the matched result, therefore the matched segment of point $P_i$ is $Argmax(x[i])$. We present the pseudo code of the inference procedure in Algorithm~\ref{infer}.

\section{Experiments}

In this section, we perform experiments on DiffMM. We first introduce the experimental setup, then evaluate the effectiveness of map matching. Furthermore, we analyze the efficiency of DiffMM during training and inference. In addition, we conduct ablation study on key modules. Finally, we evaluate the robustness of DiffMM by reducing the training data size.

\subsection{Experimental Setup}
\begin{table}
    \centering
    \caption{Dataset statistics.}
    \begin{tabular}{ccl}
         \toprule 
         Dataset & Porto & Beijing \\
         \midrule
         Number of trajectories &  1,013,437 & 1,176,097\\
         Time interval~(s)& 15 & 60\\
         Area size~($km^2$) & $11.7 \times 5.2$ & $29.6 \times 30.0$ \\
         Number of segments & 11,491 &  65,276 \\
         Number of intersections &  5,330 & 28,738 \\
         \bottomrule
    \end{tabular}
    \label{data}
\end{table}

\subsubsection{Datasets.} 
 Table \ref{data} lists the details of the two real-world trajectory data on Porto~(PT) and Beijing~(BJ), both of which are taxi trajectories. Table~\ref{data} also provides the number of segments and intersections and the area size of road networks. Since map matching in urban areas is typically more significant and difficult, we select the central urban area in Porto as the training data, which has relatively smaller area size and fewer number of road segments. To demonstrate the scalability of DiffMM, we select a region in Beijing with large area size and large number of road segments. We obtain the road networks from OpenStreetMap\footnote{\url{www.openstreetmap.org}}. 

 For every trajectory, we randomly sample the GPS points in it to generate its sparse trajectory. The average time interval of the sparse trajectory $T$ is $t/r$ where $t$ is the original interval and $r$ is a ratio in $(0, 1)$. For Porto dataset, we set $r$ to $0.2$, $0.1$, $0.05$ and $0.025$, whose average interval are 75s, 150s, 300s and 600s respectively. We set $r$ to $0.5$, $0.3$, $0.2$ and $0.1$ for Beijing dataset, the average interval are 120s, 200s, 300s and 600s respectively. For each dataset, we randomly split it into training, validating and testing sets with ratio in $40\%$, $30\%$ and $30\%$. 

\subsubsection{Baselines.} 
 To evaluate the performance of out model, we include the following methods as baselines: 1) HMM~\cite{hmm}, the most commonly-used method for map matching. 2) DeepMM~\cite{deepmm}, an end-to-end deep learning method based on Recurrent Neural Network~(RNN). 3) GraphMM~\cite{gmm}, a graph-based method which incorporate graph neural networks. 4) RNTrajRec~\cite{rntraj}, a method for trajectory recovery, we set its ratio to 1 to use it for map matching. 

\subsubsection{Implementations.} 
 In our method, we set search dist $\delta$ for trajectory encoder to 50 meters. We set the embedding dimensionality $d_{emb}$ to $128$ and the condition dimensionality $d_{cond}$ is $256$ correspondingly. We stack two layers transformer encoder with four heads for trajectory encoder. For the DiT blocks, we set the hidden dimension $d$ to $512$ and stack two layers of DiT blocks. We train the shortcut with $d\in \{1, 1/2\}$ and inference with one step~($M=1$). The learning rate is set to 1e-3. Our method is implemented in Python 3.11 with PyTorch 2.4.0. All experiments are conducted on a Linux machine with a single NVIDIA RTX 3090 GPU with 24GB memory.

\subsubsection{Evaluate Metric.} 
 We evaluate all methods' effectiveness through the accuracy of map matching. Let $R=[r_1, r_2, ..., r_l]$ be the ground-truth matched road segments of Trajectory $T$ whose length is $l$, and $\hat{R} = [\hat{r}_1, \hat{r}_2, ..., \hat{r}_l]$ be the matched road segments predicted by the model. We calculate accuracy as follows:

\begin{equation}
    Accuracy(R, \hat{R}) = \frac{1}{l} \sum_{i=1}^l \mathbf{1} \{r_i = \hat{r}_i\}.
\end{equation}

We calculate the accuracy for each trajectory and report the average accuracy over all testing trajectories. 

\subsection{Overall Performance} \label{perf}

\begin{table*}
\centering
 \caption{Evaluation of prediction accuracy. A larger value is better~(in percentage). Bold denotes the best result and underline denotes the second-best result.}
 \label{expacc}
 \begin{tabular}{c|cccc|cccc}
 \toprule
  & \multicolumn{4}{c}{Porto} & \multicolumn{4}{c}{Beijing} \\
 \cmidrule(r){2-9}
 Methods & $r=0.2$ & $r=0.1$ & $r=0.05$ & $r=0.025$ & $r=0.5$ & $r=0.3$ & $r=0.2$ & $r=0.1$ \\
 \midrule
 HMM & \underline{92.46} & \underline{83.82} & 66.62 & 40.04 & \underline{89.19} & \underline{77.51} & 68.24 & 46.46 \\
 GraphMM & 52.84 & 49.22 & 37.67 & 34.49 & 40.96 & 20.57 & 16.32 & 12.02 \\
 DeepMM & 86.38 & 83.68 & \underline{81.37} & \underline{78.69} & 76.59 & 73.19 & \underline{71.64} & \underline{68.25} \\
 RNTrajRec & 79.56 & 77.57 & 75.81 & 73.76 & 74.45 & 69.78 & 68.68 & 68.18 \\
 DiffMM~(Ours) & \textbf{93.43} & \textbf{91.47} & \textbf{89.08} & \textbf{86.87} & \textbf{90.32} & \textbf{88.45} & \textbf{87.65} & \textbf{85.39} \\
 \bottomrule
 \end{tabular}
\end{table*}

 Table \ref{expacc} shows the overall performance of models on accuracy. Numbers in bold font indicate the best performers, and underlined numbers represent the next best performers. We observe that DiffMM outperforms all baseline models on the two datasets.
 
 GraphMM has poor performance because the road graph is not constructed correctly.
 
 From the result, we have the following observations:

 \subsubsection{The sparsity of trajectory destroy the performance significantly.} As shown in Table \ref{expacc}, all methods suffer from degraded performance when the trajectory become more sparse, especially HMM, which is very sensitive to the sparsity of trajectory. For example, in Porto dataset, HMM's accuracy is $83.82\%$ when average interval is 150 seconds, however, it decreases to $40.04\%$ when average interval is 600 seconds, which decreases by 43.78 in percentage. Over all methods, DiffMM suffers least from the sparsity. 

 \subsubsection{DiffMM has significant improvement in the two datasets when the records are sparse.} Our model achieves the best accuracy on Porto and Beijing datasets, furthermore, it has remarkably significant improvement in accuracy compared to the second-best method when the GPS records are sparse. For example, in Beijing dataset, when the sample ratio is $0.1$~(average interval is 600 seconds), DiffMM is 15.28 higher in percentage than the second-best method DeepMM. 

\subsection{Efficiency Study}
\begin{table}
    \centering
    \caption{Inference and training time.}
    \begin{tabular}{ccc}
         \toprule 
         Methods & Inference~(s) & Training~(min) \\
         \midrule
         HMM & \underline{20.57} & - \\
         GraphMM & 62.79 & 26.28 \\
         DeepMM & 88.82 & \textbf{9.07} \\
         RNTrajRec & 627.65 & 868.23 \\
         DiffMM & \textbf{1.18} & \underline{10.66} \\
         \bottomrule
    \end{tabular}
    \label{eff}
\end{table}

 Table~\ref{eff} reports the average inference time per 1000 trajectories for map matching and the average training time per epoch in Beijing dataset with sample ratio $r=0.1$. Bold denotes the best result and underline denotes the second-best result. HMM does not have training time because it does not need training. As shown, our method, DiffMM, is significantly faster than other baseline models by orders of magnitude during inference. For example, DiffMM cost only 1.18 seconds per 1000 trajectories while the second-best method requires 20.57 seconds, which is about a 17-fold speedup. The training time required by DiffMM is about equal to the fastest baseline model, but DiffMM is much faster during inference and significantly outperforms the baseline model in accuracy. The efficiency of DiffMM in both inference and training demonstrates that it is efficient and practical in practice. 

\subsection{Ablation Study}

\begin{table}
    \centering
    \caption{Ablation studies on Beijing dataset by accuracy.}
    \begin{tabular}{ccccc}
         \toprule 
         Variants & $r=0.5$ & $r=0.3$ & $r=0.2$ & $r=0.1$ \\
         \midrule
         w/o Trans & 90.06 & 88.33 & 87.12 & 84.89 \\
         w/o Attn & 88.79 & 87.25 & 85.70 & 82.71 \\
         w/o Shortcut & 89.67 & 87.92 & 86.84 & 83.53 \\
         DiffMM & \textbf{90.32} & \textbf{88.45} & \textbf{87.65} & \textbf{85.39} \\
         \bottomrule
    \end{tabular}
    \label{abl}
\end{table}

 To prove the effectiveness of the key modules of DiffMM, we create three variants of it. \textbf{w/o Trans} replaces the Transformer encoder in point representation with a FFN; \textbf{w/o Attn} replaces the attention in segment representation with a simple mean calculation; \textbf{w/o Shortcut} adopts traditional diffusion instead of the shortcut. We evaluate these variants on Beijing dataset by accuracy and report the results in Table~\ref{abl}.

 We observe a drop in performance after removing the Transformer encoder of point representation, and the sparser the trajectory, the more accuracy it dropped. This demonstrates that the Transformer encoder successfully captures the context of sparse trajectory. In addition, we observe a significant drop in performance after removing the attention mechanism of segment representation, which demonstrates that it is helpful to reduce the effects of noise in trajectory. As for shortcut, it additionally considers the desire denoise step size, which is not for the traditional diffusion. This makes shortcut performs better than traditional diffusion in one step denoising. The study results in Table~\ref{abl} also demonstrate this point, shortcut obtains better results than traditional diffusion. 

\subsection{Robustness of DiffMM}

\begin{table}
    \centering
    \caption{Accuracy Varying Training Data Size.}
    \begin{tabular}{ccccc}
         \toprule 
         Training Size & $16,000$ & $32,000$ & $64,000$ & $128,000$ \\
         \midrule
         Accuracy & 86.01 & 87.91 & 89.23 & 90.03 \\
         \bottomrule
    \end{tabular}
    \label{table:gene}
\end{table}

 We evaluate the robustness of DiffMM by varying trajectory number of training data. We choose Porto dataset with sample ratio $r=0.1$ and train our model using 16,000, 32,000, 64,000 and 128,000 trajectories in the training set, respectively. We then test the model on the testing set that have 304,032 trajectories. The accuracy results are reported in Table~\ref{table:gene}. Typically, reducing the amount of training data degrades the performance of a model. DiffMM's performance does not degrade significantly while training set is shrinking, even if the training size is 16,000, it still outperforms the next best method. This shows that DiffMM extracts the features from trajectory and road network effectively.

\section{Conclusion}

 We propose \emph{DiffMM}, an encoder-diffusion-based map matching framework. It generates joint embedding of the trajectory and candidate road segments and uses it as conditioning context for one step diffusion process through shortcut model to realize map matching. Extensive experiments on large-scale trajectory datasets highlight the effectiveness and efficiency of our method compared to other methods, particularly for sparse trajectories and complex road network topologies. As for future work, it is feasible to modify the model to generate a dense sequence of road segments through the conditioned diffusion process, which is the task also known as trajectory recovery, whose target is to recover a reasonable dense trajectory for the given sparse trajectory.

\section{Acknowledgments}
 This work was partially supported by the National Natural Science Foundation of China (62472174, 62402082), the Open Research Fund of Key Laboratory of Advanced Theory and Application in Statistics and Data Science–MOE, ECNU, the Scientific and Technological Research Program of Chongqing Municipal Education Commission (KJQN202400637), and the Fundamental Research Funds for the Central Universities.

\bibliography{aaai2026}

\end{document}